\documentclass{article}

\PassOptionsToPackage{numbers}{natbib}
\usepackage[preprint]{neurips_2022}

\usepackage[utf8]{inputenc} % allow utf-8 input
\usepackage[T1]{fontenc}    % use 8-bit T1 fonts
\usepackage{hyperref}       % hyperlinks
\usepackage{url}            % simple URL typesetting
\usepackage{booktabs}       % professional-quality tables
\usepackage{amsfonts}       % blackboard math symbols
\usepackage{nicefrac}       % compact symbols for 1/2, etc.
\usepackage{microtype}      % microtypography
\usepackage{xcolor}         % colors
\usepackage{graphicx}

\usepackage{amsmath}
\title{Understanding Natural Language in Context}
\author{Avichai Levy, Erez Karpas \\
Technion - Israel Institute of Technology\\
\texttt{avichai@campus.technion.ac.il, karpase@technion.ac.il}
}

\begin{document}
\maketitle
% Use \bibliography{yourbibfile} instead or the References section will not appear in your paper
%\nobibliography{aaai22}
% My code
\begin{abstract}
Recent years have seen an increasing number of applications that have a natural language interface, either in the form of chatbots or via personal assistants such as Alexa (Amazon), Google Assistant, Siri (Apple), and Cortana (Microsoft). To use these applications, a basic dialog between the robot and the human is required.

While this kind of dialog exists today mainly within "static" robots that do not make any movement in the household space, the challenge of reasoning about the information conveyed by the environment increases significantly when dealing with robots that can move and manipulate objects in our home environment.

In this paper, we focus on cognitive robots \cite{levesque2008cognitive}, which have some knowledge-based models of the world and operate by reasoning and planning with this model. Thus, when the robot and the human communicate, there is already some formalism they can use -- the robot’s knowledge representation formalism.  

Our goal in this research is to translate natural language utterances into this robot's formalism, allowing much more complicated household tasks to be completed. We do so by combining off-the-shelf SOTA language models, planning tools, and the robot's knowledge-base for better communication. In addition, we analyze different directive types and illustrate the contribution of the world's context to the translation process.
\end{abstract}

\section{Introduction}
As time goes by, we see a significant increase in the use of virtual assistants such as Amazon's Alexa, Google Assistant, Apple's Siri, and many more. Although these virtual assistants are able to perform basic household tasks using online communication with smart home products, many more tasks remain unsatisfied. The next generation of these assistants are robots that operate in our houses, follow instructions given by us and perform much more complicated tasks. To make this dream come true, a robot must have the ability to plan a series of actions from an initial state until the required task is completed. Such a planning process could be accomplished via the PDDL formalism \cite{McDermott1998PDDLthePD}. 

Another basic requirement for household robots is the ability to reason about information that comes from the environment. This information can be visual, textual, audio, etc. As for the language reasoning part, in 2017, the Transformer neural network architecture appeared \cite{DBLP:journals/corr/VaswaniSPUJGKP17}, reaching a great breakthrough in machine translation. Later on, transformer architectures evolved, solving a larger range of NLP tasks. Yet, training an agent to perform complicated actions in the real world is still hard, expensive, and time-consuming. Therefore, a lot of recent work has been dedicated to the development of realistic virtual simulators that mimic the behavior of the real world.

\begin{figure*}
\centering
\includegraphics[width=1\textwidth]{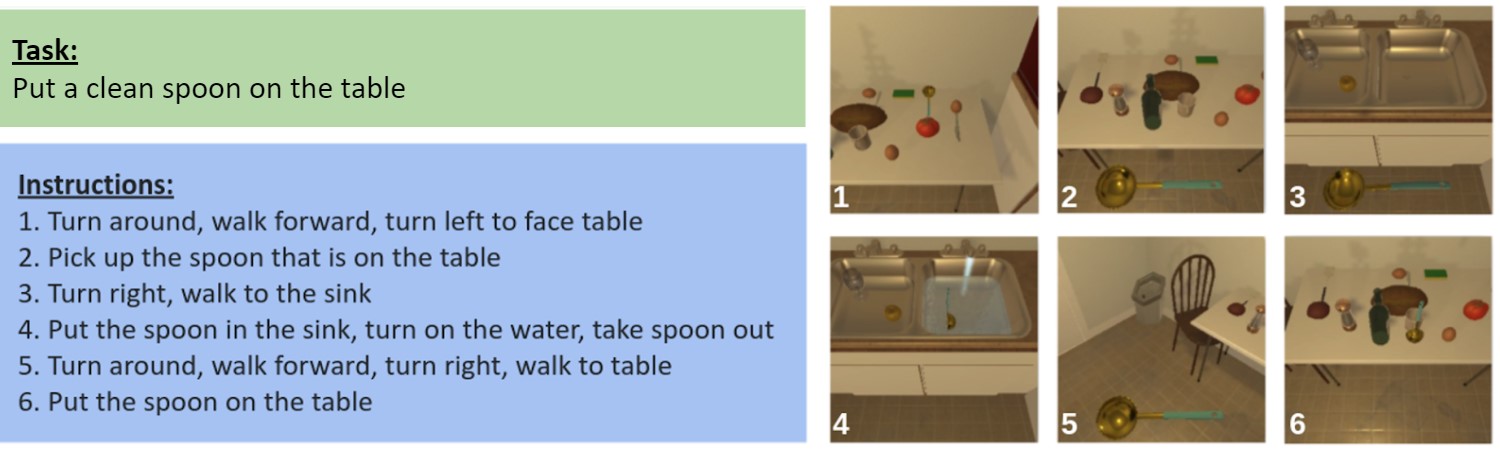}
\caption{\label{fig:ALFRED_example}An example from the ALFRED dataset. The green text box contains the high level task and the blue text box contains the high level instructions needed for accomplishing that task. The images represent the egocentric vision input of the agent at each time step. }
\end{figure*}

AI2-Thor \cite{kolve2017ai2} is an example of this kind of simulator. The AI2-Thor simulator demonstrates realistic constraints from our real world. A good example of a real-world constraint is an \textbf{irreversible state} in which some actions change the world in a way that cannot be reversed (for example: the only tomato in the scene was sliced). In 2020, \citet{shridhar2020alfred}  introduced the ALFRED (Action Learning from Realistic Environments and Directives) dataset, which is based on the AI2-Thor simulator. 
ALFRED consists of multi-modal data. It has a long sequence of instructions to achieve high-level tasks such as "Put a slice of tomato in the fridge." 
Those step-by-step instructions are combined with the egocentric vision of the robot at each time step (see \ref{fig:ALFRED_example} for example). 

Our goal in this paper is to develop a system that incorporates natural language processing and planning to enable an agent to accomplish a real-world task given in natural language. By doing so, we are able to show how changes in the language input affect the agent's capability to accomplish the task. In this work, we assume that the agent has some background knowledge about the environment it operates in (which is essential for any cognitive robot regardless of its natural language processing capabilities), such as the types of objects in the world, and their features and the possible relations between them, as well as the basic robot behaviors (modeled as actions with preconditions and effects). Moreover, we assume that at the beginning of every episode, the agent acquires complete scene information, including all the objects and their locations, using its vision tools. The last piece of the agent's input is a directive, in which a human asks the agent to perform a specific task in natural language. By combining all this information, our agent should generate a sequence of actions (in the robot's formalism) that will achieve the desired outcome of the human task. To do so, we have developed a system which combines large language models and PDDL planning. We evaluate this system on 
%and trained a model on 
the textual part of the ALFRED dataset. We show how various inputs affect the outcome of the model and suggest that the context of the environment is vital for the agent's ability to succeed as well. Our main contributions are as follows:
\begin{itemize}
    \item We developed a novel translation approach that combines natural language and planning formalism for operating agents.
    \item We show that integrating the scene context (the world's semantic information) into the model's input leads to a significant improvement in the translation process, which indicates that the context of the world is vital for an operating agent.
    \item We imply that both the Transformer's encoder and the Transformer's decoder are essential for this translation task.
\end{itemize}

\section{Related work}

Home service robots must have the ability to plan a sequence of actions to achieve their goals in the real world. This skill requires sophisticated reasoning at each time step, including interpreting multi-modal input types such as vision, language, and other sensor-type information. Thanks to environments like AI2-THOR\cite{kolve2017ai2}, Matterport 3D \cite{anderson2018vision} ,AI Habitat \cite{habitat19iccv}, and TDW\cite{gan2020threedworld}, a dramatic improvement has been made in various real-world tasks.
One of these is \textit{visual semantic planning} \cite{zhu2017visual}, which is the task of predicting a sequence of high-level actions from visual observations. The purpose of those actions, conducted by the agent, is to reach a goal state from an initial state. When addressing this kind of task, a robot operating in a human household space may need to overcome some challenges. 
For example, partially observable space or long-horizon tasks in which the decision-making at any step can depend on observations received far in the past. 
Hence, being able to properly memorize and utilize long-term history is crucial. \cite{fang2019scene}
In 2020, \citet{shridhar2020alfred} introduced the ALFRED dataset, which is based on the AI2-Thor simulator. ALFRED combines both egocentric vision and language directives to achieve everyday household tasks. Currently, the best model completed 39\% of ALFRED's tasks, which still leaves a long way to go.

Recent papers have chosen to break this problem down into separate modalities instead of solving this difficult multi-modal problem. \citet{jansen2020visually} explored this task on the ALFRED dataset, by using the GPT-2 \cite{radford2019language} language model to generate these plans from high-level task descriptions, without visual cues. In his work, \citet{jansen2020visually} showed that the GPT-2 model outperforms a baseline RNN model on this task, predicting successfully of 22.2\% actions sequences, and 53.4\% of the plans when ignoring the first action prediction in the sequence. Later on, \citet{wang2021visual} integrated a general domain knowledge graph of indoor environments with the BERT model \cite{devlin2018bert} to create better predictions, generating successfully 31.4\% of the plans. While these previous works focused on language directive translation, they do not incorporate practical planning, and therefore are not sufficient for real-world intelligent agents.

On the other hand, \citet{wang2020home} integrated Hierarchical Task Network and Probabilistic Inference to generate action sequences using multiple context types, but without natural language directives.
These papers indicate that models can achieve surprising performances using information from only a single modality.
In addition, a recent study \cite{thomason2018shifting} has found that models using input from a single modality (either vision or language) often perform nearly as well as or even better than their multi-modal counterparts.

\section{Methods}

\begin{figure*}[t]
\centering
\includegraphics[width=0.85\textwidth]{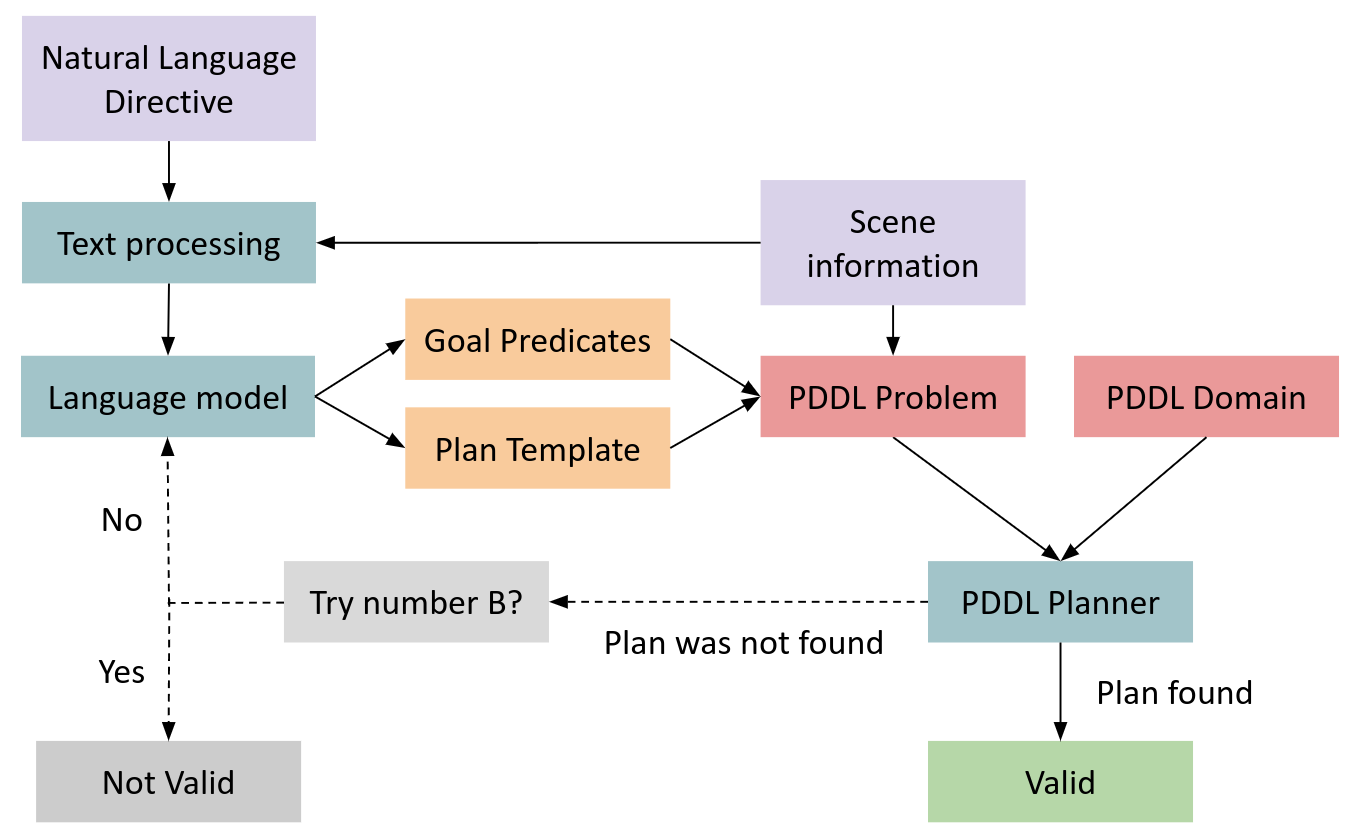} % Reduce the figure size so that it is slightly narrower than the column.
\caption{Our model architecture.}
% A natural language directive goes into a language model. After the language model generates both goal predicates and plan template, we incorporate them to create a new PDDL problem file and check if the PDDL planner is able to find a solution under the template constraints. If not, we generate another plan template and check again. This process repeats until a valid solution was found or the number of trials exceeds a constant predefined $B$.}
\label{fig_model}
\end{figure*}
As described before, we assume that our agent already has background knowledge about the available actions, objects, and predicates in the world. Another assumption is the agent has complete scene information (typically acquired using vision). Additionally, the agent is given a language directive by a human, which instructs it to perform a particular task. The agent's goal is to output a sequence of actions that   achieves the desired objective. 

In general, a human can give language commands to a robot in multiple ways. The first option is to provide the agent with high-level task description, such as "put a slice of tomato in the fridge". On the other hand, more detailed instructions could be given also, for example, "go to the kitchen, pick up the knife from the table, go to the tomato that is on the counter, slice the tomato, etc.". While both of these types exist in each sample of the ALFRED dataset, we choose to focus on the high-level task description. Moreover, in this paper we incorporate the spatial relations between the objects in the scene, which reflects the semantic context of the environment. We term the high-level task description as \textit{Task}, and the additional context as \textit{Relations}.

Furthermore, we use the task and relations to train our model to produce two  outputs: \textit{goal predicates} and \textit{plan templates}. 
The \textit{goal predicates} represent the desired outcome of a given task description. In other words, the goal predicates are the main objects' state at the end of the plan execution ("sliced tomato, tomato on counter"). On the other hand, a \textit{plan template} is the general structure of the robot's plan, which specify a sequence of actions and objects types that form the plan ("go to table, pickup knife table, go to tomato, slice tomato").

By combining these outputs together, we are able to achieve our goal, which is to translate natural language into a \textbf{valid} robot-language plan. A valid plan is a sequence of actions in a robot's language that can be performed by a household agent to achieve given tasks.

In this work, we assume that the robot's language is the Planning Domain Definition Language (PDDL, more details and background in the supplementary material).
As mentioned above, the PDDL domain contains the objects in the world and the actions' definitions (preconditions and effects). Unlike the domain file, which is predetermined, the problem file varies between each task we are trying to solve. The main components of the problem file are the task's goal predicates and the current world state, which includes the objects in the scene and their predicates.

By adding predicates to the action's parameters and the world state, we restrain the problem, allowing the PDDL planner to terminate much faster. In our research, we add three types of constraints:

\begin{itemize}
    \item \textbf{Length} - we add two predicates to the domain - $(next \ ?s_i \ ?s_j)$ and $(current\_step \ ?s_i)$.
    Each action in the domain gets the current step, $s_i$, and increases it to $s_j$, where $s_j$ is the next step number in the $(next\ s_i\ s_j)$ predicate. 
    By initializing the problem with the predicates:
    \begin{itemize}
        \item $(current\_step \ s_0)  = True$
        \item $(next \ s_i \ s_{i+1}) = True \ \ \forall i :\ 0 \leq i < T$
    \end{itemize}
     and adding the predicate $(current\_step \ s_T)$ to the goal, we force the PDDL planner to generate only plans of length $T$.
    \item \textbf{Action allowance} - for each time step $i \geq 0$, we add the predicate $(allowed\_action \ s_i)$, where \textit{action} is some action type from the domain, making the planner to create only plans that their $i$'th action type is allowed.
    \item \textbf{Objects allowance} - This predicate is similar to the previous one. The predicate is $(allowed\_object_j \ obj \ s_i)$. Which indicates if the $i$'th action in the plan, $(action\_type \ obj_1 \dots obj_k)$, may contain the object type $obj$ at location $j$. 
\end{itemize}

The combination of these constraints forces the PDDL planner to create a very specific PDDL plan. Therefore, these constraints reduce the search space, hence accelerating the process of finding a plan, if one exists. An illustration of our model's architecture is available in Figure \ref{fig_model}.
The first component is the translation unit, which is responsible for translating a natural language directive into PDDL goal and plan template. The second part of the model will combine these elements into a PDDL problem and check for a PDDL plan that solves this problem. If such a plan is not found, the model will generate another PDDL plan template and re-check for a solution. Ultimately, this process ends when a valid plan is found or the number of plans generated exceeds a constant B. We will now drill down into further details of each component.

\subsection{Language-PDDL translation}
This unit consists of two translation channels. Both channels take the same natural language directive as input. The first channel and the primary meeting point between the language part and the PDDL part is the \textit{plan templates}. A \textit{plan template} (or \textit{visual semantic plan} as in \cite{jansen2020visually}) is a sequence of actions that achieve a general goal. 
Each action consists of an action type and parameter types. These actions are not object-specific, meaning that they cannot be used together as a plan for the PDDL problem and therefore not sufficient for an operating agent either. For instance, a plan template could be "go to dining table, pick up apple dining table, go to fridge, put apple fridge". Since there might be multiple apples and tables in our scene, the robot will not know which objects it should integrate with. In other words, a valid PDDL plan must contain the objects' unique ids as well as their types.

Even though the plan templates alone are not enough to solve each problem, they are still useful. In our method, we convert these plan templates into the PDDL constraints we defined above. 
To demonstrate this, consider the following plan template:
\textit{"go to table, pick up apple table"}.

The constraint predicates that derive from this plan template are:
\begin{itemize}
    \item \textbf{Length} - since the length of the plan is 2, we add the predicates $(next\ s_0\ s_1)$, $(next\ s_1\ s_2)$, $(current\_step\ s_0)$ to the initial world state and $(current\_step\ s_2)$ to the goal predicates.
    \item \textbf{Action allowance} - for each element in the predicted sequence, we add the action allowance predicate of the element's action type and index - $(allowed\_goto \ s_0)$, $(allowed\_pickup \ s_1)$.
    \item \textbf{Objects allowance} - as in the previous case, we add the object allowance predicate for each object in the sequence. Each predicate consists of the object type, its location in the action, and its action's location in the sequence. - $(allowed\_arg_1\ table\ s_0)$,  $(allowed\_arg_1\ apple\ s_1)$, $(allowed\_arg_2\ table\ s_1)$.
\end{itemize}

By integrating these constraints into the original PDDL problem, we force the planner to produce a plan that will match our template. This restriction significantly decreases the search space, thus accelerating the search process overall. 

The second translation channel is the goal predicates unit. The goal predicates capture the user's desired outcome of a task. These predicates describe the world's state at the end of the plan's execution ("sliced tomato", "cold tomato"). In contrast to the PDDL plan, which should include specific objects, the goal predicates may be formulated in a more general way. In this paper, we use this general goal formalism rather than focusing on specific object ids. In other words, when the task's goal includes some predicates of a specific object, we accept any plan that reaches the same predicates on any instance of this object type. This is done due to the fact that there are many instances of each object type in our world, and we do not want to pick only one instance when we formulate the goal. We implement this attribute using the \textit{exists} PDDL operator.

To illustrate this, assume the PDDL goal \textit{"sliced tomato, on tomato countertop"}. This goal will be formulated as:
\textit{"(exists (?tomato0 - tomato ?countertop0 - countertop)  (and (sliced ?toamto0)  (on ?tomato0 ?countertop0)))"}

This encoding allows us to accept any plan that achieves a final world state in which there exists a sliced tomato on any countertop. After generating these two PDDL elements, we combine them with the original PDDL problem, creating a new and more restricted problem to solve. As in earlier work, we model this translation process as a sequence-to-sequence task. Moreover, in our research, we focused on two language models, GPT-2 and T5 \cite{raffel2019exploring}. We have trained separate GPT-2 models for \textit{goal predicates} prediction and for \textit{plan template} prediction. However, since training a new task on T5 requires only changing the prefix of the input, we fine-tuned a single T5 model for both tasks. 

\subsection{PDDL consistency checking}
Once a new PDDL problem has been generated from the predicted goal and the constraint  predicates, we will input both the domain of our world and the problem file into a PDDL planner. The planner will look for a valid plan that achieves our goal under the given constraints.

When the planner does find a solution, we count the plan template predictions as valid.
On the contrary, when the planner does not find a solution, we will go back to our language model and "ask" it to generate another plan template. After a new template was generated, we convert it to PDDL constraints, update the PDDL problem, and check for a valid plan that fits the new template. This re-generation of the plan template is done by taking the next prediction in the model's beam search output.
We repeat this procedure until a valid template is found or the number of templates generated exceeds a given number $B$ (in our experiment $B$ = 5).

\subsection{Various input types}
When a human approaches everyday tasks, he may have some preliminary knowledge about the world he operates in. We term this knowledge the "context of the environment" which includes, among other things, the objects in the scene, their spatial relations, and action-object pairs that commonly appear together. When providing only a task description to a robot that does not have this context knowledge, it may struggle to generate a successful plan. In this paper, we suggest that adding the context of the environment to the task description as the input for the model (rather than using the task description alone), improves the quality of the model's output. We check this hypothesis by providing our model with various input types and tracking the changes in its performance.

Concretely, we perform multiple experiments, each having its own language input. The directives that were tested are the concatenation of the high-level task ("put two bowls on the dining table") with the relations between the objects in the scene ("on tomato table, on bowl countertop, etc."). In addition, to isolate the effect of each input type, we also analyze the performance of the model when the input contains only one input type. 

\section{Evaluation}
We evaluate our model on the language part of the ALFRED dataset and show that our model is able to achieve state-of-the-art results on the \textit{visual semantic plan} generation task and \textit{valid robot-plan} generation task.

\subsection{Dataset}
The ALFRED dataset consists of 8,055 visual samples, composed of an agent's egocentric visual observations of the environment.
Each one corresponds to multiple language directives, annotated by mechanical turkers, adding up to a total number of 25,743 directives.

ALFRED has 7 different task types parameterized by 84 object classes in 120 scenes. The tasks are Pick \& Place, Stack \& Place, Pick Two \& Place, Clean \& Place, Heat \& Place, Cool \& Place and Examine in Light.
ALFRED is based on the AI2-Thor simulator, in which some actions may change the object's state in an irreversible way (a sliced potato will never be whole again). 
The evaluation data in ALFRED is divided into validation and test datasets. Each one is split also into seen and unseen environments.
The purpose of the second split is to examine how well a model generalizes to entirely unseen new spaces with novel object class variations.

\paragraph{Pre-processing}
In our work, we redivide the original ALFRED's training data into train, val, and test.
Furthermore, we combine  ALFRED's seen type validation set with our validation set and test our model both on our test data and ALFRED's validation unseen data. Since in our task we ignore the vision part of the data, we might encounter some duplicates between our datasets. Hence, we perform a cleaning process that deletes duplicate samples from the training and validation data with the same language directive as in our test datasets.

In the ALFRED dataset, there are several actions that the agent can execute. These actions are: $pick up,\ put,\ slice,\ heat,\ cool,\ clean,\ toggle$ and $go to$. By performing a single action or a sequence of actions, the agent may change the state of some objects. To track these changes, we model the state of each object by using the PDDL predicates formalism. The predicates that reflect the outcomes of these actions are $robot\_has\_obj,\ on,\ sliced,\ hot,\ cold,\ cleaned$, $toggled$, and $can\_reach$. In addition, we added another predicate, named $two\_task$ which is used as an indicator for the "pick two and place" task.

In our work, we train language models to predict both the goal predicates, which express the desired final state of each object as derived from the language directive, and the plan template, A.K.A the \textit{visual semantic plan}. To create the targets for the language models, we focused on the PDDL parameters and the high-level actions provided by the ALFRED samples. Each action in the plan template is in the form - $(action,\ arg1)$ if $action \in \{go to,\ toggle\}$, and $(action,\ arg1,\ arg2)$, otherwise. In the same way, each goal predicate is in the form - $(predicate,\ arg1, \ arg2)$, if $predicate = on$,  and $(predicate,\ arg1)$, otherwise.

\paragraph{Models input format}
Since we use two different language models in our evaluation, GPT-2 and T5, we have to adjust the input to the correct form that these models accept.
We fine-tune GPT-2 on the natural language directives and gold targets using the GPT's sos and eos tokens:
\[ 
     \mbox{$" \textless | \textit{startoftext} | \textgreater \ \textbf{directive} \ \textit{Task Type:} \ \textbf{target} \ \textless | \textit{endoftext} | \textgreater"$}\\
\]

Where  \textit{Task Type}  is either "Goal" or "Actions" according to the prediction task we are performing. 

During evaluation and testing, we feed the model with the input  "$\textless | \textit{startoftext} | \textgreater \ \textbf{directive} \ \textit{Task Type:}$",  and let it generate tokens until a $\textless | \textit{endoftext} | \textgreater$ token is generated.

On the other hand, the T5 fine-tuning process on a new task is done by providing a unique prefix before the directive.
In our work, the goal-predicates task's prefix is "translate task to goal" and the plan-template task's prefix is "translate plan to actions".
Since T5 is an encoder-decoder model, at every training step we feed the model with source and target sequences. The source phrase is the prefix with the directive, and the target phrase is either the goal predicates or the plan template.

\paragraph{Data validation}
A data sample will be considered \textit{valid} for training if its original action sequence solves the sample's problem. To check if a given solution does solve a given task, we need a PDDL domain file and a PDDL problem file. Thus, we have created a PDDL domain file using our knowledge of the objects and actions in the ALFRED world and a PDDL problem file for each sample. The ALFRED domain file consists of the rules of the ALFRED world and its object types. While the same domain file is used across all samples, the problem file is different between samples.
Moreover, creating a PDDL problem file requires knowing the world's current state, meaning, all the objects in our scene and their spatial relations. Although the ALFRED dataset samples provide some of the objects in the scene, it neither reveals all of the objects nor their spatial relations, but only their explicit coordinates in the space. To find the objects' relations, we used the initial location of each object and the scene type, taken from the ALFRED dataset, and loaded them into the AI2-Thor simulator.
By doing so, we achieved the metadata of the scene, which provides more information about objects and their spatial relations. Concretely, we create relations in the form $\textit{on} \ obj_1 \ obj_2$, where $obj_1$ is on top of $obj_2$ or inside it.
Lastly, the goal predicates for each problem were generated from the "PDDL parameters" field of every data sample.

After creating the domain and problem files, we extracted the PDDL action sequence from the "high level plan" of each sample (which specifies the objects' ids) and used the VAL plan validator \cite{1374201} to check if this plan did solve the PDDL problem of this sample. 
Samples whose gold PDDL action sequence did not achieve the goal of the problem were marked as invalid samples and were removed from the data. Eventually, the train, val, test, and test\_unseen datasets had 13893, 1650, 1010, and 682 samples, respectively. This division reflects an 80-10-10 (\%) train-val-test partition.

\paragraph{Metrics}
In our research, we implement both the evaluation measures used in \cite{jansen2020visually} and some additional accuracy measures. Moreover, we extend these measures to the \textit{goal predicate} task.
We also use the same notation of permissive scoring, which accepts predictions of objects that are similar to the original ones. ("lamp - floor lamp", "knife - butter knife"). Both tasks have per-element accuracy measures ($predicate$\textbackslash $command, \ arg_1,\ arg_2$), permissive arguments ($p\_arg1,\ p\_arg2$), full triples and full sequence accuracy measures as defined in \cite{jansen2020visually}. Notice, however, that while in the \textit{visual semantic plan} task the order of the generated text does matter ("go to table, pick up tomato table" is not the same as "pick up tomato table, go to table"), in the \textit{goal predicate} task we ignore the order of the generated predicates and measure the accuracy accordingly ("sliced tomato, cold tomato" is the same as "cold tomato, sliced tomato").

In the \textit{goal predicate} task, we also look at permissive scoring in the full predicate and sequence level. The $f\_predicate\_sim$ and $f\_seq\_sim$ measures indicate if a predicate or a sentence is wrong only in permissive objects. ("cold butter knife, hot apple" and "cold knife, hot apple" are the same) We implemented two accuracy measures for the \textit{valid robot-plan} task. The first measure is the $valid\_plans\_orig\_goal$ measure, which indicates the ratio of samples that a valid pddl plan (achieves the \textbf{original} goal predicates) was found, while following the plan template constraints. The second measure is similar to the first, except that it counts plans that achieve the \textbf{predicted} goal predicates. We term the second measure  $valid\_plans\_pred\_goal$.

% \begin{table*}[hbt!]
% \centering
% \def\arraystretch{1.1}%
% \begin{tabular}[hbt!]{|c| c|   c   c  c   c  c  |c  c|} 
% \hline
% \rule{0pt}{3ex} Model & Input & Command & Arg1 & Arg2 & F\_Action & F\_Seq \\[1mm]

% \hline
% \rule{0pt}{3ex}  
%  & Task & \textbf{0.93} & 0.75 & 0.67 & 0.63 & 0.32\\
%  GPT-2 & Relations & 0.54 & 0.14 & 0.16 & 0.10 & 0.00\\
%   &  Task + Relations & \textbf{0.93} & 0.78 & 0.74 & 0.69 & 0.46\\
%   & & & & & & \\
%   & Task & 0.91 & 0.73 & 0.63 & 0.60 & 0.29\\
%  T5 & Relations & 0.68 & 0.22 & 0.26 & 0.18 & 0.04\\
%   &  Task + Relations & 0.92 & \textbf{0.82} & \textbf{0.76} & \textbf{0.75} & \textbf{0.57}\\[1mm]
% \hline
% \end{tabular}
% \caption{\textit{Plan Template} Precision accuracy scores. Command, Arg1 and Arg2 are per-element accuracy measures. F\_Action and F\_Seq represent the ratio of correct full action triples and full actions sequence respectively. Combining the world's context lead to a significant improvement in each model, across all accuracy measures. T5 model with Task+Relation input predicts correctly 57\% of full action sequences, which is 11\% more than GPT-2.}
% \label{actions_table}
% \end{table*}

\subsection{Results}
We now analyze the results on each task, tested on two language models and three directive types.
\paragraph{Goal predicates}
 The accuracy measures in this section were calculated using the precision metric and were evaluated on ALFERD's val\_unseen dataset. The full accuracy scores on the goal predicate task are available in the supplementary material. While in the T5 model the accuracy scores were the highest on the \textit{Task + Relations} input, in the GPT-2 model the \textit{Task + Relations} input was better than the \textit{Task} input only on the full\_sequence measure. In addition, T5 outperform GPT-2 on every input type, reaching almost 90\%  accuracy across all measures and predicting correctly 85\% of exact full sequences.
These results suggest that an encoder-decoder architecture might be more suitable for goal predicate prediction. Moreover, the additional information about the environment was captured better by the T5 model than by the GPT-2 model. 
Further examples of the goal prediction of our T5 model on new and unique examples are shown in \ref{goal_examples}. These directives are different from the common tasks of ALFRED, and their intention is to check the robustness of the model. As presented in the table, the model seems to recognize well general types such as vegetables, cutlery, and baking tools.

\begin{table}[t]
\caption{Goal prediction examples}
\def\arraystretch{1.1}%
\begin{tabular}{l l}
\toprule
Input Text & Goal Predicates \\
\midrule Put a \textbf{baking tool} on the counter.  & on \textbf{spatula} pan, on pan countertop
\\\midrule
Place two \textbf{vegetables} in the drawer. & on \textbf{potato} drawer, two\_task\\\midrule
Put any type of \textbf{cutlery} on the counter.  & sliced \textbf{spoon}, on \textbf{spoon} cup, on cup countertop\\
\bottomrule
\end{tabular}
 
\label{goal_examples}
\end{table}

\paragraph{Plan template}

\begin{table}[hbt!]
\caption{\textit{Plan Template} accuracy scores.}
\centering
\def\arraystretch{1.1}%
\begin{tabular}{c c c c c c c c } 
\toprule
Model & Input & Command & Arg1 & Arg2 & F\_Action & F\_Seq & Valid Plans\\

\midrule 
 & Task & \textbf{0.93} & 0.75 & 0.67 & 0.63 & 0.32 & 0.78\\
 GPT-2 & Relations & 0.54 & 0.14 & 0.16 & 0.10 & 0.00 & 0.00\\
  &  Task + Relations & \textbf{0.93} & 0.78 & 0.74 & 0.69 & 0.46 & 0.89\\
\midrule
  & Task & 0.91 & 0.73 & 0.63 & 0.60 & 0.29 & 0.72\\
 T5 & Relations & 0.68 & 0.22 & 0.26 & 0.18 & 0.04 & 0.13\\
  &  Task + Relations & 0.92 & \textbf{0.82} & \textbf{0.76} & \textbf{0.75} & \textbf{0.57} & \textbf{0.91}\\
\bottomrule
\end{tabular}
\label{actions_table}
\end{table}

\begin{table}[hbt!]
\caption{Full actions triples accuracy per 8 action types in ALFRED.}
\begin{center}
\begin{tabular}[hbt!]{c c c c c c c c c c c} 
\toprule
Model & {GoTo} & {Pickup} & {Put} & {Cool} & {Heat} & {Clean} & {Slice} & {Toggle} & {Avg.}\\
\midrule
GPT$_T$ & 68 & 40 & 68 &  \textbf{85} & 82 & \textbf{78} & 39 & 75 & 67 \\
T5$_T$ & 66 & 36 & 66 & 79 & 83 & 74 & 47 & 77 & 66\\
GPT$_{TR}$ & 75 & 56 & 67 & 78 & 80 & 75 & \textbf{55} & 75 & 70 \\
T5$_{TR}$ & \textbf{79} & \textbf{65} & \textbf{72} & 83 & \textbf{84} & \textbf{78} & \textbf{55} & \textbf{97} & \textbf{77} \\
\bottomrule
\end{tabular}
\label{per_action_accuracyy}
\end{center}
\end{table}

Table \ref{actions_table} contains the models' scores on the \textit{plan template} task.
In contrast to the \textit{goal predicate} task results, both models achieve the highest score on the Task + Relations input.
On the Task-only input, GPT-2 predicts correctly 32\% of full original action sequences, which is better result from previous work (22\%), but might be due to the training dataset changes. Furthermore, when adding the scene context to the model's input, T5 outperforms GPT-2 across almost all measures, predicting correctly 57\% of the target plans in comparison to GPT's 46\%.
These results outperform recent work \cite{jansen2020visually} on \textit{visual semantic plan} generation from natural language directives, which was also trained and evaluated on the ALFRED dataset. Moreover, the actions triples accuracy per each action type are presented in table \ref{per_action_accuracyy}, where the subscript \textit{T} and \textit{TR} refers to Task and Task+Relations inputs respectively.

% \NewDocumentCommand{\rot}{O{45} O{1em} m}{\makebox[#2][l]{\rotatebox{#1}{#3}}}%

\paragraph{Valid robot plan}
The models' scores on the \textit{valid robot plan} task on the original goal predicates are presented in table \ref{actions_table} under the \textit{Valid Plans} column. As shown in the table,
T5 model generates valid plans for 91\% of the samples, where 57\% of them are the exact target plans. That is, 34\% of the plan templates that T5 predicted were not the same as the original plans, but eventually achieved the desired goal.

In addition, when the input is non-informative of the required task, such as the objects' relations alone, GPT-2 is not able to generate any valid plan. On the other hand, T5 generated valid plans for 59\% of the samples with respect to the predicted goal, but only 13\% plans with respect to the original goal predicates. These results suggest that T5 might generate easier goal predicates when the input is non-informative rather than succeeding to predict valid plans. 

Lastly, GPT-2 achieve better results on the original goals, in contrast to T5 which succeed more on the predicted goal predicates. This difference might be due to the fact that T5 succeed more on the \textit{goal predicate} task than GPT-2. The full table with the \textit{Valid Plans} scores on the predicted goal predicates is provided in the supplementary material.

\begin{figure}[hbt!]
\centering
\includegraphics[width=1\textwidth]{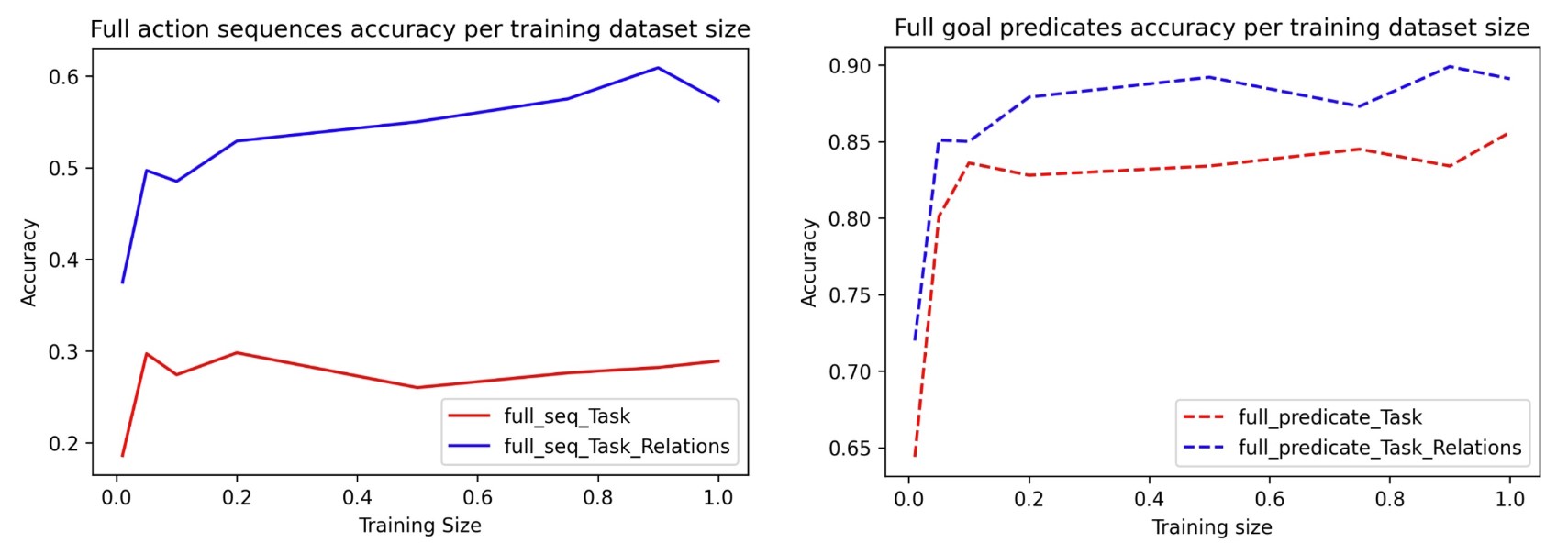} % Reduce the figure size so that it is slightly narrower than the column.
\caption{Few-shot accuracy of full action sequences and full goal predicates.}
\label{few_shot_fig}
\end{figure}

\paragraph{Few shot learning}
In our setting, creating data samples for training is time-consuming and expensive. Hence, the ability of a model to perform successful few-shot learning is crucial. To evaluate this capability, we have created multiple training sets by downsampling the original data into smaller fractions and trained different T5 models on each set. As shown in figure \ref{few_shot_fig}, we see that with only 5\% of the training data, our models are able to predict actions sequences and goal predicates nearly as well as models that were trained on the full dataset. These results suggest that our model is able to perform successful few-shot learning. We are planning to test this assumption in other domains in future work.

\section{Conclusions and future work}
In this work, we have developed a novel approach for translating natural language to plans that accomplish everyday household tasks. While previous work \cite{jansen2020visually} reached 53.4\% accuracy on action generation by training GPT2 model on task-only input \textbf{and ignoring the first action generated by the model}, our model predicts precisely 57\% of the ALFRED dataset plans without ignoring the first action in the predicted sequence, and generates valid plans for another 34\% of the samples, reaching a total number of valid plans for 91\% of unseen environments tasks. Furthermore, we show that our model performs successful few-shot learning. These plans were generated by using only natural language data for task description and scene information. Our approach combines both language models and PDDL tools, working together as a whole to generate a valid PDDL plan that will achieve the language directive goal. our contribution reflects \textbf{two ways of improvements} - (1) we utilize not just the task description, but also extra information about the environment's context, and (2) we employ the more powerful T5 model, which is better able to exploit this extra information.

%(1) we choose the T5 model, which is more powerful and predicts better actions sequences than GPT2, even on the task-only input, and (2) improve both models by providing them with 

Looking forward, there are some future directions we would like to investigate. One of them is generating valid PDDL plans for object-specific goals. This task is more challenging since its goal state requires particular objects' predicates to be changed rather than any object of this type (for example, warming a \textbf{green} cup instead of any cup).
Another challenge is defining a more conservative PDDL problem and domain that avoids changing basic attributes of objects (for example, put a \textbf{whole} tomato on the table). In addition, since the ALFRED dataset contains limited number of objects and only seven types of our everyday tasks, achieved by performing only eight action types, which is much less than the set of all possible actions that an household agent should be able to perform, we would like to improve our zero-shot predictions on new and unseen object types, and check the few-shot learning capabilities of our model on other domains. we will release our code and data under an open source licence once the paper is published.

\bibliography{main}

% \documentclass{article}

% % if you need to pass options to natbib, use, e.g.:
% %     \PassOptionsToPackage{numbers, compress}{natbib}
% % before loading neurips_2022

% % ready for submission
% \usepackage{neurips_2022}
% \bibliographystyle{ieeetr}

% to compile a preprint version, e.g., for submission to arXiv, add add the
% [preprint] option:
%     \usepackage[preprint]{neurips_2022}

% to compile a camera-ready version, add the [final] option, e.g.:
%     \usepackage[final]{neurips_2022}

% to avoid loading the natbib package, add option nonatbib:
%    \usepackage[nonatbib]{neurips_2022}

% \usepackage[utf8]{inputenc} % allow utf-8 input
% \usepackage[T1]{fontenc}    % use 8-bit T1 fonts
% \usepackage{hyperref}       % hyperlinks
% \usepackage{url}            % simple URL typesetting
% \usepackage{booktabs}       % professional-quality tables
% \usepackage{amsfonts}       % blackboard math symbols
% \usepackage{nicefrac}       % compact symbols for 1/2, etc.
% \usepackage{microtype}      % microtypography
% \usepackage{xcolor}         % colors

% \title{Formatting Instructions For NeurIPS 2022}

% \begin{document}

\section{Appendix} 

\paragraph{PDDL} 
\label{PDDL}
The Planning Domain Definition Language (PDDL) \cite{McDermott1998PDDLthePD} is a language family that allows us to define a planning problem. PDDL is an action-centred language, inspired by the well-known STRIPS formulations of planning problems \cite{fikes1971strips}.
Mathematically, a STRIPS instance is a quadruple $\langle F,A,I,G\rangle$, in which each component has the following meaning:
\begin{itemize}
    \item \textit{F} -- a set of facts describing the possible states of the
world, $2^F$.
    \item \textit{A} -- a set of actions. Each action $a \in A$ consists of a set of preconditions $pre(a)$, add effects $add(a)$, and delete effects $del(a)$. Applying $a$ is possible in a state $s$ where $pre(a) \subseteq s$, and results in the state $s[\langle a \rangle] = (s\setminus del(a)) \cup add(a)$.
    \item $\textit{I} \subseteq F$ is the initial state of the world.
    \item $G$ -- the goal of the problem. The goal $G$ is a set of facts $G \in F$. A state $s$ satisfies a goal if $G \subseteq s$.
\end{itemize}

A plan $\pi$ is a sequence of actions. $\pi = \langle a_0, a_1, \dots , a_n\rangle$ is applicable from state $s_0$ if $a_0$ is applicable at $s_0$ and $\langle a_1, \dots , a_n\rangle$ is applicable from $s_1 := s_0[\langle a_0 \rangle]$.
We denote the state reached by following plan $\pi$ from state $s$ by $s[\pi]$, and say that a plan $\pi$ achieves a goal $G$ if $G \subseteq s[\pi]$. 
The PDDL language generalizes the STRIPS setting into domain description and problem description.

The domain description contains the definitions of object types, predicates, as well as the actions' preconditions and effects. These elements are the aspects that do not change regardless of what specific situation we are trying to solve. On the other hand, the PDDL problem description is more specific and defines exactly what objects exist in the scene, what their current states are, and what the goal is. For example, let us assume that the world contains apples, tomatoes, cucumbers, knives, and tables in three different colors: blue, yellow, and green. The actions that could be conducted on an object are: pickup, put, and slice. A problem file can be described as follows: The current scene contains two apples, three tomatoes, two knives, a yellow table, and a blue table. In the initial state, all the objects (besides the blue table) are on the yellow table. The goal is to put a slice of an apple on the blue table. Many PDDL tools have been developed over the years. One major part is PDDL planners, which read PDDL files (domain and problem) and use them to find a sequence of actions that solves the problem.
Another tool is a plan validator \cite{1374201}, which checks if a given plan solves a specific PDDL problem.

\paragraph{Experiment details}
We used the medium version of GPT2 model and fine-tuned it for 25 epochs with learning rate of 5e-5, 100 warm-up steps and 0.01 weight decay. The batch sizes for the Task, Relations, and Task+Relations inputs are 32, 8, and 4 respectively. For the T5 model, we fine-tuned the base version of this model, also for 25 epochs. The learning rate used to tune this model is 1e-4. The batch sizes for the Task, Relations, and Task+Relations inputs are 32, 16, and 16 respectively. All other hyper-parameters which are not mentioned here were set to their default values. Both models were optimized using the Adam optimizer.
All experiments were conducted using the NVIDIA A100 GPU, with 40GB GPU RAM. Tuning the models on the Task, Relations, and Task+Relations inputs types last for 2, 2, and 3 hours on the T5-base model, and  1, 4, and 5 hours on the GPT2-medium model, respectively.
\subsection{Results}

\paragraph{Goal predicates}

In this section, we provide the full results of the \textit{Goal Predicate} task. Table \ref{goal_table_appendix} include all metrics and reported scores, divided to two language models and three input directives. The scores in Table \ref{goal_table_appendix} are divided into two categories: strict and permissive. Predicate, Arg1, and Arg2 are per-element accuracy measures. F\_Predicate and F\_Seq represent the ratio of correct full predicate pairs (or triples when the predicate type is $on$) and full predicate sequences, respectively. As shown in the table, on both models, the best results are achieved with the task description and the objects' relations as input. In addition, T5 outperforms GPT-2 on every input type, achieving almost 90\% accuracy on every measure. Table \ref{goal_examples_appendix} contains new and unseen directives types, and their corresponding predicted goal that our T5 model (which was trained on the Task + Relations input) generates. In the left column are the new task inputs for the model. Each task was paired with the objects' relations in the scene. These directives are different from the common tasks of ALFRED, and their intention is to check the robustness of the model.

\begin{table}[t]
\caption{\textit{Goal Predicates} Precision accuracy scores.}
\centering
\def\arraystretch{1.1}%

\begin{tabular}{c c c c c c c c} 
\toprule
Scoring Type & Model & Input & Predicate & Arg1 & Arg2 & F\_Predicate & F\_Seq\\

\midrule  
 &  & Task & 0.80 & 0.77 & 0.79 & 0.76 & 0.66\\
& GPT-2 & Relations & 0.02 & 0.01 & 0.02 & 0.02 & 0.01\\
&  &  Task + Relations & 0.73 & 0.71 & 0.81 & 0.70 & 0.74\\
Strict & & & & & & & \\
 &  & Task & 0.89 & 0.86 & 0.84 & 0.85 & 0.78\\
& T5 & Relations & 0.09 & 0.08 & 0.05 & 0.08 & 0.04\\
&  &  Task + Relations & \textbf{0.92} & \textbf{0.89} & \textbf{0.88} & \textbf{0.89} & \textbf{0.85}\\

\midrule
&  & Task & 0.80 & 0.81 & \textbf{0.89} & 0.80 & 0.72\\
&  GPT-2 & Relations & 0.02 & 0.01 & 0.02 & 0.02 & 0.01\\
&  &  Task + Relations & 0.73 & 0.74 & 0.83 & 0.73 & 0.77 \\
Permissive & & & & & & & \\
 &  & Task & 0.89 & 0.89 & 0.85 & 0.89 & 0.84\\
& T5 & Relations & 0.09 & 0.09 & 0.05 & 0.09 & 0.04\\
&  &  Task + Relations & \textbf{0.92} & \textbf{0.92} & \textbf{0.89} & \textbf{0.92} & \textbf{0.88}\\
\bottomrule
\\
\end{tabular}
\label{goal_table_appendix}
\end{table}

\begin{table}[t]
\caption{Goal prediction examples}
\def\arraystretch{1.1}%
\begin{tabular}{l l}
\toprule
Input Text & Goal Predicates \\
\midrule
Put either tomato or potato or lettuce \\on the counter. & on lettuce countertop, cold lettuce \\
\midrule
Put either tomato or potato on the counter, \\\textbf{avoid using lettuce}.  & on \textbf{potato} countertop, cold \textbf{potato}\\
\midrule Put a baking tool on the counter.  & on spatula pan, on pan countertop
\\\midrule
Place two vegetables in the drawer. & on potato drawer, two\_task\\\midrule
Put any type of cutlery on the counter.  & sliced spoon, on spoon cup, on cup countertop\\\midrule

Put some element in the fridge.  & sliced potato, on potato fridge, cleaned potato \\
\bottomrule
\end{tabular}
 
\label{goal_examples_appendix}
\end{table}

\begin{table}
\caption{\textit{Valid Robot Plan} accuracy scores.}
\centering
\def\arraystretch{1.1}%
\begin{tabular}{c c c c} 
\toprule
Model & Input & Valid\_Plan\_Orig\_Goal & Valid\_Plan\_Pred\_Goal \\

\midrule

 & Task & 0.78 & 0.69\\
 GPT-2 & Relations & 0.00 & 0.00\\
  &  Task + Relations & 0.89 & 0.83\\
  \midrule
  & Task & 0.72 & 0.77\\
 T5 & Relations & 0.13 & 0.59\\
  &  Task + Relations & \textbf{0.91} & \textbf{0.97}\\
\bottomrule
\end{tabular}

\label{valid_robot_plan}
\end{table}

\paragraph{Valid robot plans} Table \ref{valid_robot_plan} contains the scores on the valid plans task. The
Valid\_Plan\_Orig\_Goal score is the ratio of valid plans that achieve the \textbf{original} goal, while the Valid\_Plan\_Pred\_Goal score reflects a similar ratio, but focuses on the \textbf{predicted} goal predicates. While GPT-2 achieves better results on the original goal predicates, T5 predicts more accurately the valid plans in the opposite case. This difference may be due to the fact that GPT-2 struggles to predict valid goal predicates in contrast to T5.

% \end{document}

\end{document}